\documentclass[10pt,twocolumn,letterpaper]{article}

\usepackage{cvpr}

\usepackage[utf8]{inputenc}
\usepackage{times}
\usepackage{epsfig}
\usepackage{booktabs}           
\usepackage{graphicx}
\usepackage{amsmath}
\usepackage{amssymb}
\usepackage{physics}      
\usepackage{xparse}       
\usepackage[font=small,bf]{caption}   
\usepackage{subcaption}   
\usepackage{xcolor}
\usepackage{cite}         
\usepackage{enumitem}     
\usepackage{microtype}    
\usepackage{xstring}          

\definecolor{citecolor}{RGB}{34,139,34}  
\usepackage[pagebackref=true,breaklinks=true,letterpaper=true,colorlinks, citecolor=citecolor,bookmarks=false]{hyperref}

\makeatletter
\renewcommand{\paragraph}{%
  \@startsection{paragraph}{4}%
  {\z@}{1.25ex \@plus 1ex \@minus .2ex}{-1em}%
  {\normalfont\normalsize\bfseries}%
}
\makeatother

\newcommand{\lidar}{{LiDAR}}
\renewcommand{\deg}{{$^{\circ}$}}
\DeclareMathOperator*{\argmin}{\arg\!\min}
\DeclareMathOperator*{\argmax}{\arg\!\max}

\renewcommand{\vec}[1]{\mathbf{#1}} 

\newcommand{\bp}{\mathbf{p}}

\newcommand{\mytilde}{\raise.17ex\hbox{$\scriptstyle\sim$}}

\newcommand{\xcoarse}{\tilde{\vec{x}}_\text{coarse}}
\newcommand{\xfine}{\tilde{\vec{x}}_\text{fine}}

\newcommand{\best}[1]{\textbf{#1}}
\newcommand{\second}[1]{\underline{#1}}

\definecolor{john_color}{RGB}{117,112,179} 
\definecolor{shenlong_color}{RGB}{27,158,119} 
\definecolor{raquel_color}{RGB}{217,95,2}     
\definecolor{andrei_color}{RGB}{231,41,138}   
\definecolor{julieta_color}{RGB}{102,166,30}  

\newif\ifarxiv
\arxivtrue

\cvprfinalcopy 

\def\cvprPaperID{7332} 

\ifcvprfinal
\fi
\begin{document}

\title{Deep Multi-Task Learning for Joint Localization, Perception, and Prediction}

\author{
  John Phillips$^{1,2}$ \quad Julieta Martinez$^{1}$ \quad Ioan Andrei B{\^{a}}rsan$^{1,3}$ \\
  Sergio Casas$^{1,3}$ \quad  Abbas Sadat$^{1}$ \quad Raquel Urtasun$^{1,3}$ \\ \\
  $^{1}$Uber Advanced Technologies Group \quad \quad $^{2}$University of Waterloo \quad \quad $^{3}$University of Toronto\\
{\tt\small \{john.phillips,julieta,andreib,sergio.casas,asadat,urtasun\}@uber.com}
}

\maketitle

\ifarxiv
  \pagestyle{plain}
  \thispagestyle{plain}
\fi

\begin{abstract}

Over the last few years, we have witnessed tremendous progress on many subtasks of autonomous driving
including perception, motion forecasting, and motion planning. 
However, these systems often assume that the car is accurately localized against a high-definition map.
In this paper we question this assumption,
and investigate the issues that arise in state-of-the-art autonomy stacks under localization error.
Based on our observations, we design a system that jointly performs perception, prediction, and localization.
Our architecture is able to reuse computation between the three tasks,
and is thus able to correct localization errors efficiently.
We show experiments on a large-scale autonomy dataset, demonstrating the efficiency
and accuracy of our proposed approach.

\end{abstract}

\section{Introduction}
\label{sec:intro}

Many tasks in robotics can be broken down into a series of subproblems that are easier to study
in isolation, and facilitate the interpretability of system failures~\cite{zhou2019action}.
In particular, it is common to subdivide the self-driving problem into five critical subtasks:
(i) Localization: placing the car on a high-definition (HD) map with centimetre-level accuracy.
(ii)  Perception: estimating the number and location of dynamic objects in the scene.
(iii) Prediction: forecasting the trajectories and actions that the observed dynamic objects might do in the next few seconds.
(iv) Motion planning: coming up with a desired trajectory for the ego-vehicle, and
(v) Control: using the actuators (\ie, steering, brakes, throttle, \etc) to execute the planned motion.

\begin{figure*}[!ht]
	\includegraphics[width=0.90\linewidth,trim=0mm 0mm 4mm 0mm,clip=true]{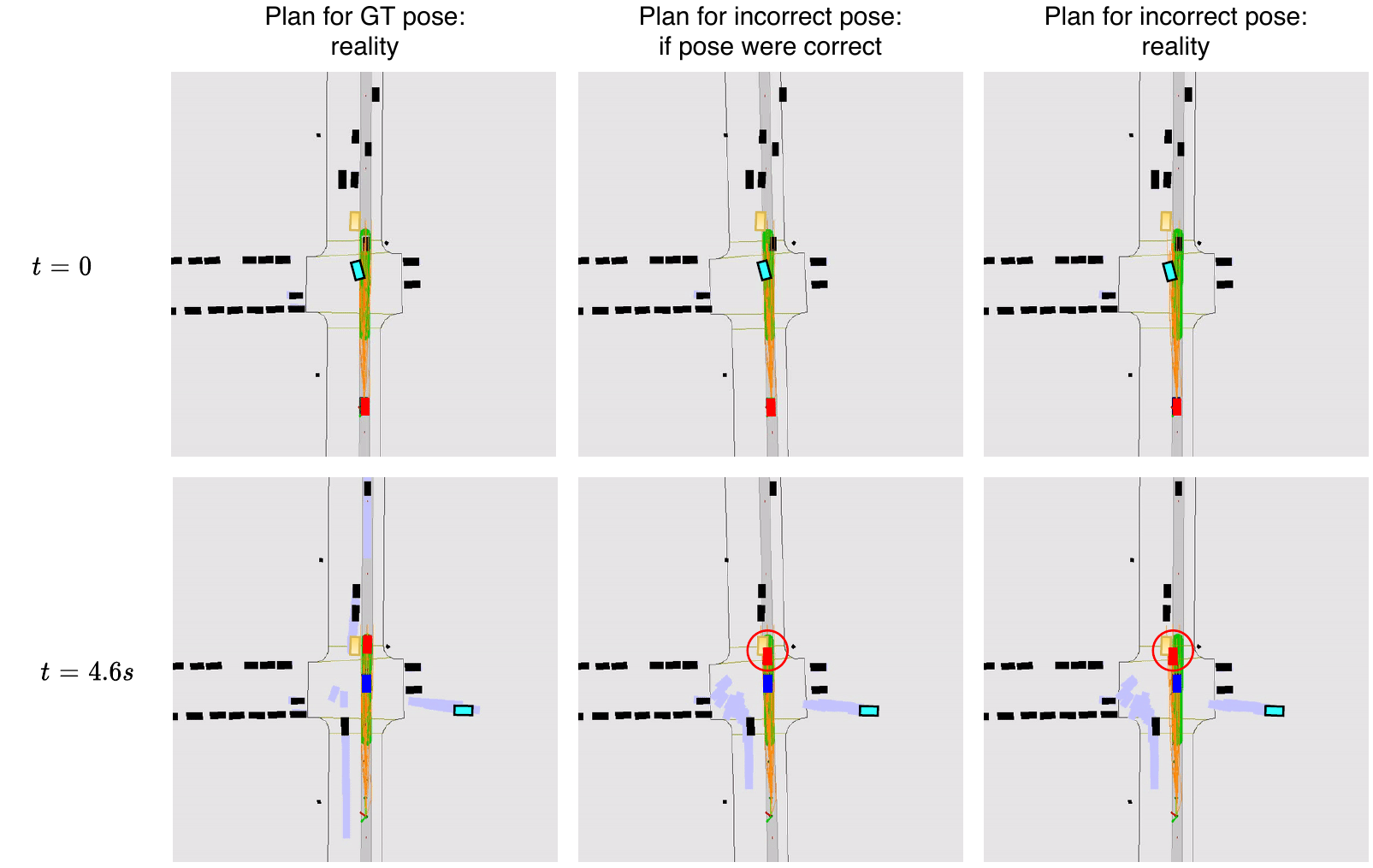}
	\caption{
		{\bf A scenario where a small amount of localization error results in a collision.}
		The top row visualizes the first time step, and the bottom row
		visualizes a later time step where a collision occurs.
    GT labels are black rectangles, and the pale blue rectangles are forecasted object trajectories.
    The SDV is the red rectangle, with its GT trajectory in a dark blue rectangle.
    The samples predicted by the motion planner are shown as orange lines.
    The 3 columns visualize different variants of the same scenario.
		(Left)
		The planned trajectory of the SDV when there is no localization error.
		(Middle) What the SDV “thinks” is happening, based on its estimated pose that has error ($x$, $y$, yaw) =
		(10 cm, 0 cm, 1.5 deg).
		(Right) What the SDV is actually doing when
		subject to the pose error; this is the same trajectory as shown in
		the middle image, but rigidly transformed so that the initial pose
    agrees with the GT pose.
    The collision (red circle) occurs because the yellow vehicle
    is not perceived at $t=0$ due to occlusion (by the cyan vehicle); the
    localization error then causes the SDV to go into the
		lane of opposite traffic which results in a collision.
}
\label{fig:motivation}
\vspace{-0.5em}
\end{figure*}

Moreover, it is common to solve the above problems \emph{sequentially}, such that the output of one
sub-system is passed as input to the next, and the procedure is 
repeated iteratively over time.
This \emph{classical} approach lets researchers focus on well-defined problems that can be studied independently,
and these areas tend to have well-understood metrics that measure progress on their respective sub-fields.
For simplicity, researchers typically study autonomy subproblems under the assumption that its inputs are correct.
For example, state-of-the-art perception-prediction (P2) and motion planning (MP) systems
often take HD maps as input, thereby assuming access to accurate online localization.
We focus our attention on this assumption and
begin by studying the effect of localization errors on modern autonomy pipelines.
Here, we observe that
localization errors can have serious consequences for P2 and MP systems, resulting in
missed detections and prediction errors, as well as bad planning that leads to larger
discrepancies with human trajectories, and increased collision rates.
Please refer to Fig.~\ref{fig:motivation} for an example of an autonomy error caused
by inaccurate localization.

In contrast to the classical formulation,
recent systems have been designed to perform multiple autonomy tasks jointly.
This \emph{joint} formulation often comes with a shared neural backbone that decreases computational
and system complexity, while still producing interpretable outputs that make it easier to diagnose system failures.
However, these approaches have so far been limited to jointly
performing perception and prediction (P2)~\cite{fast-and-furious,intentnet,ilvm,pnpnet},
P2 and motion planning (P3)~\cite{neural-motion-planner,dsdnet,p3}, 
semantic segmentation and localization~\cite{radwan2018vlocnet++} or  
road segmentation and object detection~\cite{teichmann2018multinet}.

In this paper, and informed by our analysis of the effects of localization error,
we apply the joint design philosophy to the tasks of
localization, perception, and prediction; we refer to this joint setting as
\emph{LP2}.
We design an LP2 system that shares computation between the tasks,
which makes it possible to perform localization with as 
little as $2$ ms of computational overhead
while still producing intepretable localization and P2 outputs.
We evaluate our proposed system on a large-scale dataset in terms of motion planning metrics,
and show that the proposed approach matches the performance of a traditional system with separate localization and perception components,
while being able to correct localization errors online, and having reduced run time and engineering complexity.

\section{Related Work}
\label{sec:related}


We provide a brief overview of existing approaches for the tasks that we study
(perception, prediction, and localization), followed by a discussion
of common multi-task paradigms for deep learning, and a review of recent work
in characterizing and addressing the system-level challenges in perception.


\paragraph{Object Detection and Motion Prediction:}
Detecting actors and predicting their future motion from 
sensor data is one of the fundamental tasks in autonomous driving. 
While object detection and motion
forecasting can be modeled as independent
tasks~\cite{zheng2016generating,pixor,cui2019multimodal,shi2019pointrcnn,chai2019multipath,vectornet,phan2020covernet},
models that jointly perform both tasks~\cite{fast-and-furious,intentnet,pnpnet,zhang2020stinet} 
have been shown to provide a number of benefits, such as fast inference, 
uncertainty propagation, and overall improved performance.


\paragraph{Localization:}
The objective of localization is to accurately and precisely determine the
position of the ego-vehicle with respect to a pre-built map. 
Localization methods can be based on a wide variety of sensors, such as
differential GPS in the form of
Real-Time Kinematic systems~\cite{wan2018robust,joubert2020developments},
\lidar{}~\cite{levinson2007map,yoneda2014lidar,nagy2018real,deep-gill,lu2019deepvcp},
cameras~\cite{jegou2011aggregating,kendall2015posenet,sarlin2019coarse}, 
RADAR~\cite{barnes2019masking,tang2020rsl} or
combinations of such sensors~\cite{wolcott2014visual,ma2019exploiting,zhou2020da4ad}.
While purely geometric algorithms for \lidar{} localization such as iterative
closest-point~\cite{yoneda2014lidar} have been shown to
be effective,
recent work
has shown that learned 
representations~\cite{deep-gill,lu2019deepvcp,zhou2020da4ad,du2020dh3d, sarlin2020superglue} can 
lead to improved robustness and scalability.

\paragraph{Multi-Task Learning:}
Compared to end-to-end approaches for autonomous agents that learn to directly
map sensor readings to control
output~\cite{bojarski2016end,bansal2018chauffeurnet,kendall2019learning}, 
multi-task
modular approaches have been shown to perform better
empirically~\cite{zhou2019action}, while also
being more interpretable thanks to human-readable intermediary representations
like semantic segmentation~\cite{zhou2019action}, 
object detections~\cite{dsdnet}, occupancy forecasts~\cite{p3} and planning cost
maps~\cite{neural-motion-planner}. Furthermore, Liang~\etal~\cite{liang2019multi} have
shown the benefits of jointly performing mapping, object detection, and optical
flow from \lidar{} and camera data.

The wide range of recent multi-task learning approaches can be divided into
two major areas. One line of work is focused primarily on developing and
understanding network
architectures, such as those
leveraging 
a common backbone with task-specific
heads~\cite{collobert2008unified,kaiser2017one,kendall2018multi,teichmann2018multinet,fast-and-furious,
sarlin2019coarse,du2020dh3d,cao2020unifying}, 
cascaded approaches where some tasks rely on the outputs of
others~\cite{dai2016instance,hashimoto2017joint,he2017mask,neural-motion-planner,dsdnet}, 
or
\textit{cross-talk} networks such as Cross-Stitch~\cite{misra2016cross}
which have completely separate per-task networks,
but share activation information.

%
%
Modular learning approaches such as Modular
Meta-Learning~\cite{alet2018modular},
aim to construct reusable modular architectures
which can be re-combined to solve new tasks.
Side-Tuning~\cite{side-tuning} proposes an incremental approach where new tasks
are added to existing neural networks in the form of additive side-modules that
are easy to train, and have the advantage of leaving the weights of the
original network unchanged, bypassing issues such as catastrophic forgetting.

Another line of work is concerned with the optimization process itself. 
The most straightforward approach is sub-task weighting, which may be
based on uncertainty scores~\cite{kendall2018multi}, learning
speed~\cite{chen2018gradnorm} or performance~\cite{guo2018dynamic}. 
Other methods have explored multi-task learning
by performing multi-objective optimization explicitly~\cite{sener2018multi}, 
by regularizing task-specific networks through soft parameter 
sharing~\cite{yang2017trace}, 
or through knowledge distillation~\cite{bucilua2006model,clark2019bam}.
Please see Crawshaw~\cite{crawshaw2020multi} for a detailed survey of
multi-task deep learning.

\paragraph{Planning Under Pose Uncertainty:}
  The task of planning robust trajectories under pose uncertainty has been
  studied in the past, with previous methods formulating it as a continuous
  POMDP which can be solved with an iterative linear-quadratic-Gaussian
  method~\cite{van2012motion}, or as an optimal control problem solved using
  model-predictive control~\cite{indelman2015planning}.
More recently, Artu{\~{n}}edo~\etal~\cite{artunedo2020motion} focus on autonomous vehicles and
incorporate the pose uncertainty in a probabilistic map representation
that is then leveraged by a sampling-based planner, 
while Zhang and Scaramuzza~\cite{zhang2020fisher} propose an efficient way of
estimating visual localization accuracy for use in motion planning.
\ifarxiv
Finally, Amini~\etal{}~\cite{amini2019variational} achieve robust planning in
the face of coarse topometric maps and inaccurate localization by learning to
predict distributions over control commands using a variational formulation.
\fi
However, none of these
approaches model other dynamic actors and the uncertainty in their own motion,
and they do not study the complex interplay between pose uncertainty and
state-of-the-art perception systems.

\paragraph{System-Level Analysis:}
A number of recent papers have studied the correlations between task-level
metrics, such as object detection, and system
performance~\cite{philion2020learning}. 
This line of work has
shown that while task-level metrics serve as good predictors of overall system
performance, they are unable to differentiate between similar errors that can
however lead to very different system behaviors. 
A related line of work analyzed the impact of sensor and inference
latency on object detection in images~\cite{li2020towards} and
\lidar{}~\cite{han2020streaming,frossard2020strobe}. 
At the same time, the simultaneous localization and mapping (SLAM)
community~\cite{nardi2015introducing,davison2018futuremapping,bujanca2019slambench} has recently proposed
extending SLAM evaluation beyond trajectory accuracy~\cite{kitti},
 towards system-level metrics like
latency, power usage, and computational costs.


\section{The Effects of Localization Error}
\label{sec:effect}

\begin{figure*}[!ht]
  \vspace{-0.75em}
	\centering
	\includegraphics[width=0.95\linewidth,trim=0mm 0mm 0mm 0mm,clip=true]{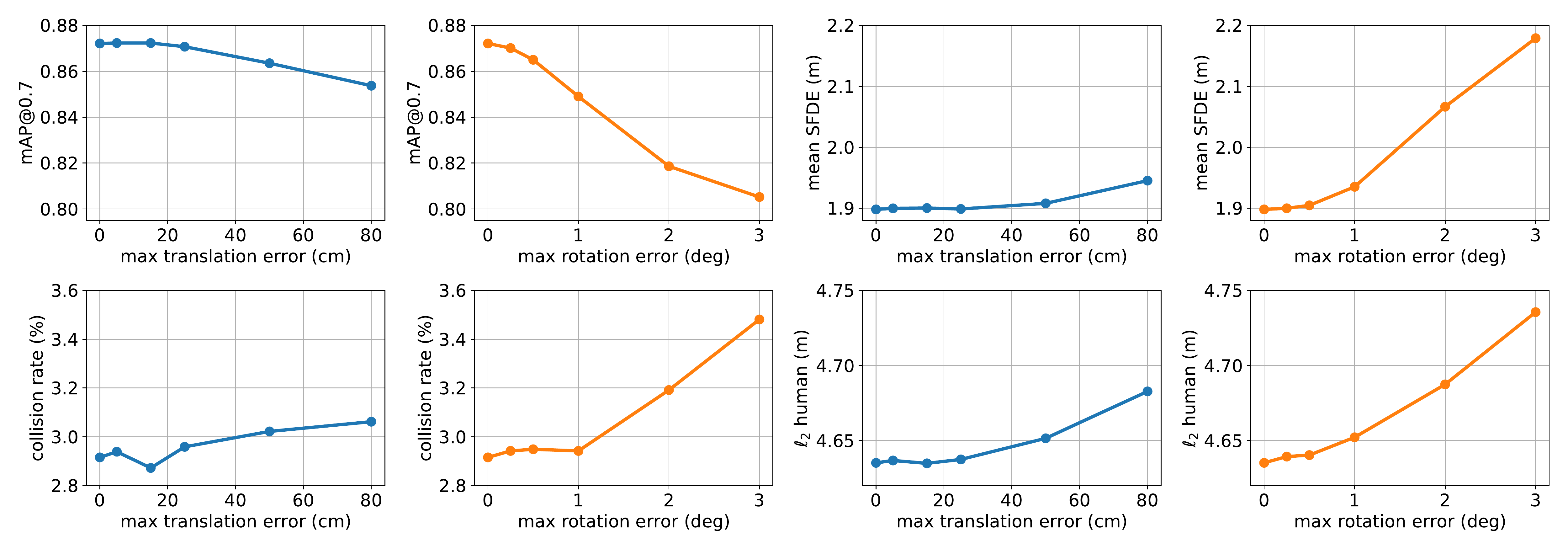}
  \vspace{-0.5em}
	\caption{%
		{\bf The effects of localization error on perception-prediction and motion planning.}
		(Top)
		The effects of perturbing the ego-pose on P2.
		SFDE is the mean displacement error across all samples
		at the 5s mark as defined in \cite{ilvm}, and mAP@0.7 is the mean
		average precision evaluated at an IOU of 0.7.
		(Bottom)
		The effects of perturbing the ego-pose on planning.
		Collision rate is the percentage of examples 
    for which the planned path
		collides with another vehicle or pedestrian within the 5s simulation, and $\ell_2$ human is the
		distance between the planned path and the ground truth human-driven path at the 5s mark.}
	\label{fig:jitter}
  \vspace{-0.5em}
\end{figure*}

Since state-of-the-art perception-prediction (P2) and motion planning (MP)
stacks make extensive use of accurate localization on high-definition maps
(often assuming perfect localization~\cite{intentnet,ilvm,bansal2018chauffeurnet,neural-motion-planner}),
we study the effects of localization error on
a state-of-the-art P2 and MP pipeline.
We begin by describing how these modules work and how they use localization.

\paragraph{Perception-Prediction (P2):}
P2 models are tasked with perceiving actors and predicting their future trajectories
to ensure that motion planning has access to safety-critical information about the scene for the entire duration of the planning horizon.
We study the state-of-the-art Implicit Latent Variable Model~\cite{ilvm} (ILVM),
the latest of a family of methods that use deep neural networks with voxelized LiDAR inputs to jointly
perform detection and prediction~\cite{fast-and-furious, intentnet, neural-motion-planner}.  
ILVM encodes the whole scene in a latent random variable and uses a deterministic decoder
to efficiently sample multiple scene-consistent trajectories for all the actors in the scene.
Besides LiDAR, the ILVM backbone takes as input a multi-channel image with semantic aspects
of the rasterized map (\eg, one channel encodes walkways, another encodes
lanes, and so on, for a total of 13 layers~\cite{ilvm}),
which the model is expected to use to improve detection and forecasting.
While the LiDAR scans are always processed in the vehicle frame,
the scans and the map are aligned using the pose of the car.
Thus, localization error results in a \emph{misalignment} between the semantic map and the LiDAR scan.

\paragraph{Motion Planning (MP):}
Given a map and a set of dynamic agents and their future behaviours, the task of the motion planner
is to provide a route that is safe, comfortable, and physically realizable to the control module.
We study the state-of-the-art Path Lateral Time (PLT) motion planner~\cite{plt}, a 
method that samples physically realizable trajectories, 
evaluates them, and selects the one with the minimal cost.

The PLT planner receives $S\!=\!50$ Monte Carlo samples from the joint distribution over the trajectories of 
\emph{all} actors $\{Y^1, \dots, Y^S\}$ from the P2 module.
It then samples a small set of trajectories $\tau \in \mathcal{T}(\mathcal{M}, \mathcal{R}, \mathbf{x}_0)$ 
given the map $\mathcal{M}$, high-level routing $\mathcal{R}$ and the current state of the SDV $\mathbf{x}_0$.
The planned trajectory
{\small
$   
\tau^* = \argmin_{\tau \in \mathcal{T}(\mathcal{M}, \mathcal{R}, \mathbf{x}_0)} \sum_{s=1}^S c(\tau, Y^s)
$
}
is then computed to be the one with the minimum expected cost over the predicted futures, as defined by
a cost function $c$ that takes into account safety and comfort.
In this case, bad localization gives the planner a wrong idea about the layout of the static parts of the scene.


\subsection{Experimental setup}%
\label{ssec:m1-exp-setup}

\begin{figure*}[t]
	\centering
	\includegraphics[width=0.80\linewidth,trim=10mm 125mm 102mm 10mm,clip=true]{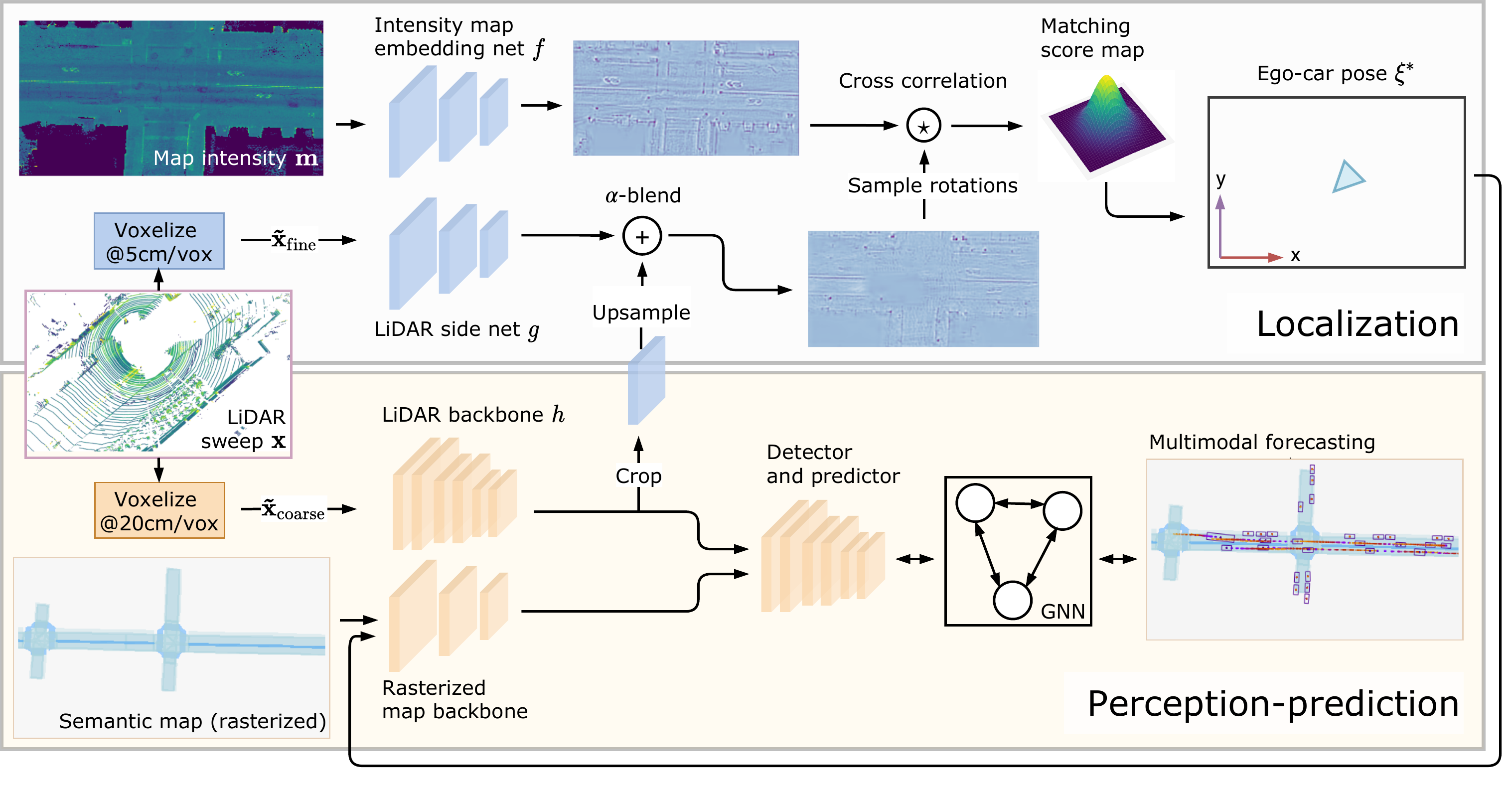}
  \vspace{-0.9em}
	\caption{The architecture of the combined localization and perception-prediction (LP2) model.}
	\label{fig:architecture}
  \vspace{-1.2em}
\end{figure*}

\paragraph{LP3 Dataset:}
Evaluating the localization, perception, prediction, and motion planning
tasks requires a dataset that contains accurately localized
self-driving segments, together with the corresponding HD appearance maps
(to evaluate localization), as well as annotations of dynamic objects in the scene, their
tracks, and their future trajectories (to evaluate P3).
To the best of our knowledge, no current public dataset satisfies all these criteria.\footnote{A few days after the submission deadline, the nuScenes dataset~\cite{caesar2020nuscenes} added support for an appearance layer, thereby enabling similar experiments to the ones we present in this work.}
Therefore, we use our own LP3 dataset.
The LP3 dataset is named after the ability it provides to evaluate
Localization (L) as well as Perception, Prediction, and motion Planning (P3). LP3
is a subset of the ATG4D dataset~\cite{ilvm,neural-motion-planner,intentnet},
that also has appearance maps available.
In particular, our maps have 2d images of aggregated \lidar{} intensity that summarize the appearance of the ground (\cf the top left of Fig.~\ref{fig:architecture}),
and are between 6 and 12 months old by the time the SDV traverses the scene.
The dataset is comprised of 1858 sequences of 25 seconds each, all captured in a large North American city.

Besides bounding boxes for vehicles, pedestrians and bicycles in the scene,
the dataset provides \emph{semantic map} annotations, such as
lanes, traffic signs and sidewalks.
Importantly, LP3 also provides a map \emph{appearance layer}
comprised of the LiDAR intensity of the static elements of the scene
as captured by multiple passes of LiDAR scans through the area
(please refer to the top left of Figure~\ref{fig:architecture} for an example).
Our LP3 dataset makes it possible to evaluate methods that jointly perform appearance-based \lidar{} localization and P2,
and to quantify motion planning metrics.

\paragraph{Simulating Localization Error:}
We simulate localization error and study its effects on downstream autonomy tasks.
Given a maximum amount of noise $m \in \mathbb{R}$ (which we call \emph{maximum jitter}),
we perturb the ground truth pose on evaluation frames
by sampling translational or rotational noise from a uniform distribution $\varepsilon \sim \mathcal{U}\left(-m, m \right)$.
To understand the effects of different types of noise,
we evaluate translational noise and rotational noise independently.

\paragraph{Metrics:}
For perception, we focus on the mean average precision metric with at least 70\% overlap between the
predicted and the ground truth boxes (mAP@0.7)~\cite{intentnet}.
For prediction, we report the mean scene final displacement error (mean
SFDE\footnote{We use meanSFDE instead of minSFDE because for large
  numbers of samples ($S=50$ in our case), minSFDE is overly optimistic.
  The presence of unrealistic false positive trajectory samples can interfere
  with the SDV, causing it to break or swerve, creating a dangerous situation.
  However, the minSFDE will not capture this dangerous behavior as long as there is
  at least one good sample. Please refer to~\cite{casas2020importance} for a more
  detailed discussion.
})
between the ground truth and the predicted trajectory after 5 seconds 
(\ie, planning horizon)~\cite{ilvm}.

We run the planner at the beginning of the
segments, and let the trajectory unfold for 5 seconds.
We then measure the percent of segments for which there is a collision,
and the $\ell_2$ distance between the predicted trajectory and the trajectory followed by the human driver after 5 seconds.

Note that in our setting all the actors are ``passive'', in the sense that they follow
their pre-recorded trajectory independently of the actions taken by the SDV.
This is often called an \emph{open-loop} evaluation.
While evaluating our task on a \emph{closed-loop} setting would be more desirable,
building a simulator of reactive agents and counterfactual sensor inputs comes with its own set of challenges
(\eg, realistic \lidar{} simulation, realistic controllable actors, \etc) and
is out of the scope of our work.

\paragraph{Results:}
We show the effects of perturbing the pose of the ego-vehicle on P2
in the top row of Figure~\ref{fig:jitter}.
We observe that the performance of ILVM is barely affected by translational jitter up to 25 cm,
and rotational jitter up to 0.5\deg.
Larger amounts of translational noise have little effect (\mytilde 2\% mAP, .05 mean SFDE) up to 80 cm,
while the effect is stronger for rotational error (\mytilde 7\% mAP, .30
mean SFDE) up to 3\deg.

We show the effects of perturbing the ego-vehicle pose on motion planning in the bottom row of Figure~\ref{fig:jitter}.
Similar to P2, MP performance does not degrade much until there is translational
noise above 25 cm (or rotational noise above 0.5\deg).
We also observe that large translation errors have small effects relative to
rotational noise for both collision rate and distance to human route.
While this is somewhat expected (as rotational error can cause straight paths to run into sidewalks
or incoming traffic), it is interesting to formally quantify these effects.


\section{Joint Localization, Perception, \& Prediction}
\label{sec:method}

We now formulate a model that performs joint localization and P2 (LP2).
First, we lay out the key challenges that we would like our system to overcome,
and then explain our design choices in detail.

\subsection{System Desiderata}

\paragraph{Low Latency:}
In order to provide a safe ride,
a self-driving car must react quickly to changes in its environment.
In practice, this means that we must minimize the time from perception to action.
In a na{\"i}ve, cascaded autonomy system, the running time of each component adds up linearly,
which may result in unacceptable latency.

To reduce the latency of the localization system,
it is common to use Bayesian filtering to provide high-frequency pose updates.
In this case, a belief about the pose is maintained over time and updated through the
continuous integration of different levels of evidence from wheel autoencoders, IMUs, or camera and LiDAR sensing.
In this context, the external sensing step
(\eg, carried out via iterative closest point alignment between the LiDAR reading and an HD map)
typically carries the strongest evidence, but is also the most expensive part of the system.
Therefore, it is critical to keep the latency of the sensing step of the localization filter low.

\paragraph{Learning-Based Localization:}
Localization systems with learned components are typically better
at discerning semantic aspects of the scene that are traditionally
hard to discriminate with purely geometric features
(\eg, growing vegetation, tree stumps, and dynamic objects),
and have the potential of being more invariant to appearance changes due to 
season, weather, and illumination~\cite{pit30m}.
Therefore, we would like to incorporate a learning-based localization component
in our system.
Moreover, since P2 systems are typically heavily driven by learning,
it should be possible to incorporate learning-based localization
by sharing computation between the two modules, resulting in
reduced overhead to the overall LP2 system.

\paragraph{Simple Training and Deployment:}
We would like our joint LP2 system to be easier to train and deploy than its classical counterpart.
Given the large amounts of ML infrastructure invested around P2 systems
(\eg, on dataset curation, labelling, active learning, and monitoring),
it makes sense to design a localization subsystem that can be trained as a smaller addition
to a larger P2 model~\cite{sculley2015debt}.
This should also make it easier to iterate on the more lightweight localization module
without the need to retrain the more computationally-expensive P2 component.


\subsection{Designing an LP2 System}

We now explain our model design choices, highlighting the ways they overcome
the aforementioned challenges and achieve our design goals.
We show an overview of the proposed architecture in Figure~\ref{fig:architecture}.

\paragraph{Input Representation:}
Our system receives \lidar{} as input, which is then converted to a
bird's-eye view (BEV) voxelization with the channels of the 2D input corresponding to the
height dimension~\cite{pixor}.
Despite P2 and localization models both relying on some form of voxelized \lidar{} input,
perception-prediction models often use a coarser \lidar{} resolution
(\eg, 20cm~\cite{fast-and-furious,neural-motion-planner}) to accomodate larger regions, 
while matching-based localizers typically require a finer-grained resolution to localize with higher
precision~\cite{deep-gill}.

Using only a fine resolution voxelization for an LP2 model
would be simplest, but imposes large run-time efficiency costs.
Therefore, to accommodate these resolution differences, our method
simultaneously rasterizes the incoming \lidar{} point cloud $\vec{x}$ into
two tensors of different resolutions, $\xcoarse$ for perception and $\xfine$
for localization.

\paragraph{Perception and Prediction:}
For our P2 subsystem, we rely on ILVM~\cite{ilvm}, whose robustness to localization
error we quantified in Section~\ref{sec:effect} -- this corresponds to the lower part
of Figure~\ref{fig:architecture}.
The proposed P2 approach contains four main submodules. (i) A lightweight network processes a rasterized
semantic map centred at the current vehicle pose. We pass our estimated pose to this module.
(ii) Another neural backbone $h$ extracts features from a coarsely voxelized 
\lidar{} sweep $\xcoarse$. 
These two features maps are concatenated and passed to (iii) a detector-predictor
that encodes the scene into a latent variable $Z$,
and (iv) a graph neural network where each node represents a detected actor,
and which deterministically decodes samples from $Z$
into samples of the joint posterior distribution over all actor trajectories.

\paragraph{Localization:}
We approach localization using 
ground
intensity localization with deep \lidar{} embeddings~\cite{deep-gill}.
%
The idea behind ground intensity localization~\cite{levinson2007map} is to align
the (sparse) observed \lidar{} sweep $\vec{x}$
with a pre-built (dense) map of the \lidar{} intensity patterns of the static scene, $\vec{m}$.
This localizer learns deep functions that produce spatial embeddings
of both the map $f(\vec{m})$ and \lidar{} sweep $g(\xfine)$ before alignment.
Following existing work~\cite{deep-gill} we parameterize the vehicle pose using
three degrees of freedom (DoF), $x$, $y$, and yaw, represented as $\xi \in
\mathbb{R}^3$.

Given a small set of pre-defined translational and rotational offsets,
we compute the dot product between the transformed sweep and the map embeddings,
and choose the pose candidate $\xi^*$ from a pre-defined set (near the original
pose estimate)
with the highest correlation as the maximum-likelihood estimate of the vehicle
pose:
\begin{align}
  \label{eq:gill-match}
  \xi^* = \argmax_{\xi} 
	 \pi\!\left(g(\tilde{\vec{x}}_{\text{fine}}\right), \xi) \cdot f(\vec{m}) 
  \triangleq \argmax_\xi \bp(\xi)
\end{align}
where $\pi$ is a function that warps its first argument based on the 3-DoF
offset $\xi$, and $\cdot$ represents the dot product operator. In practice,
this matching is done more efficiently by observing that the
dot products can be computed in parallel with respect to the translational
portion of the pose candidates by using a larger-region $\vec{m}$
and performing a single cross-correlation rather than multiple dot products for
each DoF in the rotation dimension. 
\ifarxiv
Each cross-correlation can be
  performed very efficiently in the Fourier domain to allow real-time
operation~\cite{deep-gill}.
\fi


\paragraph{Multi-Resolution Feature Sharing:}
An important advantage of localizing using \lidar{} matching is that in contrast 
to, \eg, point cloud-based
localizers~\cite{du2020dh3d}, it uses the same BEV input representation as P2,
enabling a substantial amount of computation to be shared between both systems.
However, as discussed earlier, the inputs to the P2 and localization backbones
use different resolutions, which can make information fusion difficult.
We address this issue
by upsampling a crop of the \lidar{} feature map computed by the coarse
perception backbone to match the resolution of the finer features in the
localization backbone. We then add the feature maps together using a weighted
sum to produce the final localization embedding, as depicted in
Fig.~\ref{fig:architecture}.
This allows localization \lidar{} embeddings to be computed with very
little run-time or memory overhead compared to the base perception-prediction network.

\subsection{Learning}

We optimize the full model using side-tuning~\cite{side-tuning}.
We first train the heavier perception-prediction module,
and then add the \lidar{} branch of the localizer as a side-tuned module.
In the second stage, we freeze the weights of the perception-prediction network 
(yellow modules in Fig.~\ref{fig:architecture}),
and only learn the map and online branches of the localizer (purple modules in Fig~\ref{fig:architecture}).
There are three benefits to this approach:
first, there is no risk of catastrophic forgetting in the perception-prediction task,
which can be problematic as it typically requires 3--5$\times{}$ more
computation to train than the localizer alone;
second, we do not need to balance the loss terms of the localization vs.\
perception-prediction, eliminating the need for an additional hyperparameter; 
third, training the localizer network can be done much faster than the full
system, since the P2 header no longer needs to be evaluated and fewer
gradients needs to be stored.

\paragraph{Perception and Prediction:}
We train the P2 component using supervised learning by minimizing a loss which
combines object detection with motion forecasting, while accounting for the
multimodal nature of the trajectory predictions. The P2 loss is therefore
structured as
\begin{equation}
  \mathcal{L}_\textsc{P2} = \mathcal{L}_\textsc{Det} + \alpha
  \mathcal{L}_\textsc{Pred},
\end{equation}
where
$\mathcal{L}_\textsc{Det}$ optimizes a binary cross entropy term for the object
detections and one based on smooth-$\ell_1$ for the box regression
parameters~\cite{pixor},
$\mathcal{L}_\textsc{Pred}$ optimizes the ELBO of the log-likelihood of the
inferred $n$ trajectories over $t$ time steps, conditioned on the
input~\cite{ilvm}, and $\alpha$ represents a scalar weighting term selected empirically.
%

\paragraph{Localization:}
Learning $f$ and $g$ end-to-end
produces representations that are invariant to \lidar{} intensity calibration,
and ignore aspects of the scene irrelevant to localization. Learning
is performed by treating localization as a classification task and
minimizing the cross-entropy between the discrete distribution of $\bp(\xi)$
and the ground truth pose offset $\bp^{\text{GT}}$ expressed using one-hot
encoding~\cite{deep-gill}:
\begin{equation} \label{eq:gill-loss}
  \mathcal{L}_{\textsc{Loc}} = -\sum_\xi \bp(\xi)^\text{GT} \log\bp(\xi).
\end{equation}

  The online and map embedding networks $f$ and $g$ use an architecture based
  on the P2 map raster backbone and do not share weights. In
  Section~\ref{sec:experiments} we also show that it is possibly to
  significantly improve run time by down-sizing $g$ while keeping $f$ fixed
  with little impact on overall performance.
  We refer to this architecture as a Pixor backbone~\cite{pixor}.


\section{Experiments}%
\label{sec:experiments}

We design our experiments to test the accuracy of our multi-task model on the 
joint localization and P2 (LP2) task.
We also show how these improvements translate to
safe and comfortable rides based on motion planning metrics.
We refer to the task of doing localization, P2, and motion planning as LP3.


\paragraph{Dataset and Metrics:}
We use the LP3 dataset (\cf Sec.~\ref{sec:effect}), for all our experiments.
To evaluate localization accuracy, following prior work~\cite{wei2019learning}, 
we report the percentage of frames on which 
the localizer matches the ground
truth exactly, and where it matches the ground truth or a neighbouring offset as
recall @ 1 (r@1) and recall @ 2 (r@2) respectively. In our setting, the former metric
corresponds to exactly matching the ground truth, up to our state space
resolution (5cm
and 0.5$^\circ$), while the latter corresponds to being inside a
15cm$\times{}$15cm$\times{}$1.5$^\circ$ region centered at the ground truth. 
%
For P2, we focus on mAP@0.7 for detection and mean SFDE for prediction, 
as in Sec.~\ref{sec:effect}.
For motion planning, besides collision rate and $\ell_2$ distance to human trajectory (see Sec.~\ref{sec:effect}),
we also measure lateral acceleration, jerk, and progress towards the planning goal.



\paragraph{Experimental Setup:}
There are multiple ways to design an experiment that tests a localizer.
One alternative is to start the state estimation at the identity and later align
the produced trajectory with the ground truth (as is often done in SLAM~\cite{slam}).
Alternatively, online localization often initializes the robot pose at the ground truth location,
and measures how far a localizer can travel before obtaining an incorrect 
pose~\cite{deep-gill,ma2019exploiting,zhou2020da4ad}.
By definition, these setups assume that the initial pose is correct, and do 
not test the ability to recover
from localization failure -- which is crucial for self-driving.
Instead, we assume a self-driving scenario where the localization of the pose
is initially incorrect.
As such, we perturb the true pose of the vehicle 
and thus measure
the ability of the localizer to recover from this failure,
as well as the ability of P2 and MP to deal with localization failure.
The perturbations are performed following the same uniform noise policy
described in Section~\ref{ssec:m1-exp-setup} with 0.5 metres for translation
and 1.5 degrees for rotation.







\begin{table*}
  \centering
  \small
\addtolength{\tabcolsep}{-2.5pt} 
\begin{tabular}{lccrrrrrrr}
    \toprule
    Model & P2 pose & Planning Pose & r@1 $\uparrow$ & r@2 $\uparrow$ 
          & Collision $\downarrow$ & $\ell_2$ human $\downarrow$ & Lat.\ acc.
    $\downarrow$ & Jerk $\downarrow$ & Progress  $\uparrow$ \\
    & (GT, N, L) & (GT, N, L) & (\%) & (\%) & (\% up to 5s) & (m @ 5s) & (m/s$^2$) & (m/s$^3$) & (m @ 5s) \\
    \midrule
    ILVM & GT & GT  & - & - & \best{2.915} & \best{4.64} & \best{2.13} & \best{1.82}
         & \best{24.95} \\
    ILVM & GT & N   & - & - & 3.168 & \second{4.68} & 2.21 & \second{1.83}
         & \best{24.95} \\
    ILVM & N  & N   & - & - & 3.511 & 4.70 & \second{2.20} & \second{1.83}
         & \second{24.92} \\
    \midrule
    Joint LP2 -- Ours (Tiny Pixor) & N & N & \second{46.6}  & \second{93.5} & 2.962
                               & \best{4.64} & \best{2.13} & \best{1.82}
                               & \best{24.96} \\
    Joint LP2 -- Ours (Big Pixor) & N & N & \best{52.5}    & \best{96.9}
                               & \second{2.922} & \best{4.64} & \best{2.13}
                               & \best{1.82} & \best{24.95} \\
    \bottomrule
\end{tabular}
  \vspace{-0.1em}
  \caption{
      {\bf Motion planning evaluation using pose estimate and actor predictions.}
      For the P2 and Planning poses: GT denotes ground truth (the pose was not altered);
      N denotes that localization noise was added (translation and rotation
      sampled uniformly at random from [-0.5m,+0.5m] and [-1.5\deg,+1.5\deg],
      respectively).
      Big Pixor refers to the largest width Pixor Embedding Net from Fig~\ref{fig:matching-perf-runtime}, and Tiny
      Pixor refers to the smallest. Bold denotes the best results (within an
      epsilon threshold) and
      underlines second best results.
  }
  \label{tab:motion-planning}
  \addtolength{\tabcolsep}{2.5pt} 
  \vspace{-0.3em}
\end{table*}

\begin{figure}
  \centering
  \vspace{-0.2em}
  \includegraphics[width=1\linewidth,trim=5mm 5mm 5mm 5mm,clip=true]{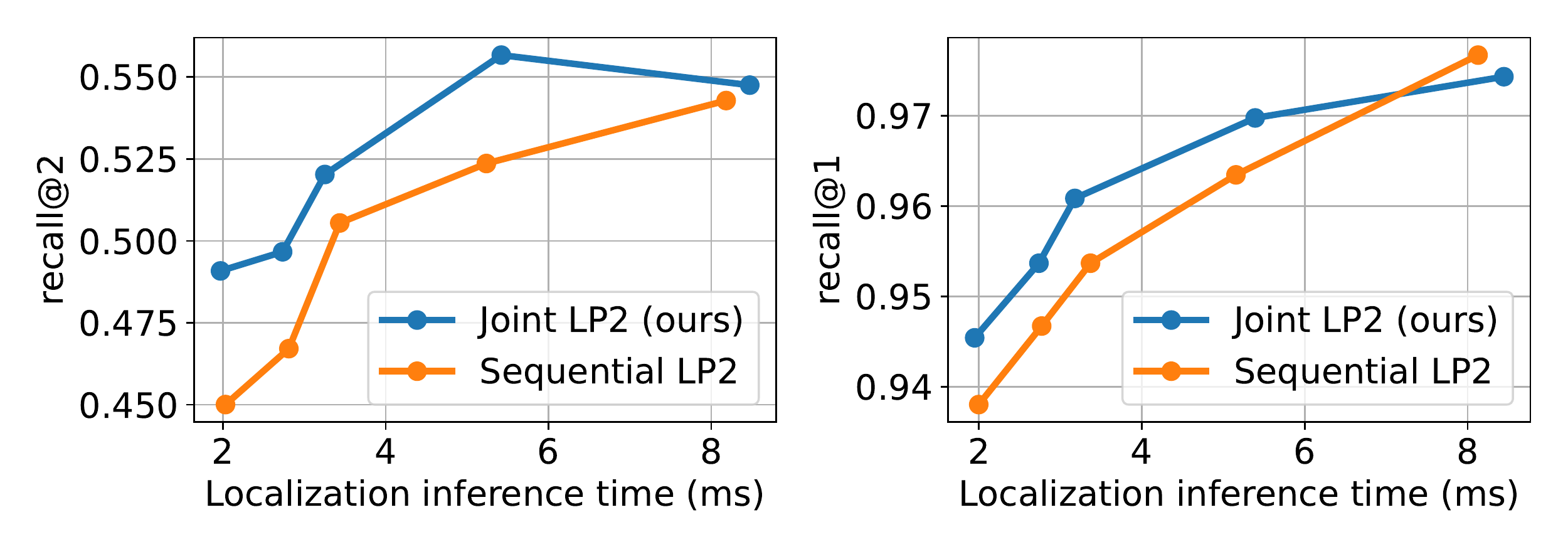}
  \caption{\textbf{Localizer embedding runtime vs.\ recall.} 
      The localization performance and runtime of the single-task (\ie, sequential)
      and multi-task (\ie, joint) localization methods.
      Faster inference is achieved by narrower and shallower networks
      for the online \lidar{} embedding.\label{fig:matching-perf-runtime}}
  \vspace{-0.5em}
\end{figure}

\begin{table}
  \centering
  \small
\begin{tabular}{lrrr}
    \toprule
    Model & Time (ms) & r@1 & r@2 \\
    \midrule
    \lidar{} Localizer~\cite{deep-gill} & 25.92 & 0.52 & 0.95 \\
    \lidar{} Localizer (Pixor-based) &  2.79 & 0.47 & 0.95 \\ 
    Joint LP2 (Ours) &        1.95 & 0.49 & 0.95 \\
    \bottomrule
\end{tabular}

  \vspace{-0.20em}
  \caption{
    \textbf{Localization inference time comparison.} While being nearly
    identical in terms of matching accuracy when comparing models with recall
    @ 2 performance similar to~\cite{deep-gill},
    the proposed approach is much faster, due to a more efficient architecture and sharing computation with
    the perception backbone.
  }
  \label{tab:deep-gill-baselines-vs-multi-task-comparison}
  \vspace{-1.00em}
\end{table}

\paragraph{Implementation Details:}
We train our model for 5 epochs using the Adam~\cite{adam} optimizer using 16 GPUs.
The coarse \lidar{} tensor $\xcoarse$ is rasterized at 20cm/voxel, 
while the fine tensor $\xfine$ uses
5cm/voxel. The spatial region corresponding to the coarse \lidar{} voxelization
is 144m$\times$80m$\times$3.2m, while the spatial region corresponding to the fine \lidar{}
voxelization is 48.05m$\times$24.05m$\times$3.2m.
Reducing the spatial extent of the high-resolution rasterization
reduces the run time of the system without sacrificing performance.
The localization search range covers $\pm$0.5m relative to the initial vehicle pose
estimate in the $x$ and $y$ dimension, discretized at 5cm intervals, and 
[-1.5$^\circ$, -1.0$^\circ$, -0.5$^\circ$, 0.0$^\circ$, 0.5$^\circ$,
1.0$^\circ$, 1.5$^\circ$] in the yaw dimension. We do not use any height
information from the maps, which are encoded as BEV images. When
training the localizer, we add uniform noise to the ground truth pose, up to
the size of the search range.

\paragraph{Results:}
In Table~\ref{tab:motion-planning}, we evaluate localization and P2
performance through motion planning metrics. Notably, despite significant
variation in localization metrics, both of our localization models perform
similarly well when evaluated in terms of the motion planning metrics.
These results further confirm the observations from our jitter experiments (Figure~\ref{fig:jitter}):
both P2 
and a short-term rollout of PLT perform similarly well when
subject to a modest amount of localization error.
This means that
besides localization accuracy (which is important from an interpretability perspective),
we have plenty of room to optimize for latency and simplicity when designing the
localization component of an LP2 architecture.
Our tiny Pixor-based localizer only takes 2ms of overhead on top of the P2 subsystem,
while providing a robust learned localization signal to the autonomy system.

\paragraph{Ablation Study:}
We perform an ablation study to investigate the
trade-off between matching performance and inference time in the localization part of our system.
We show our results in Fig.~\ref{fig:matching-perf-runtime}.

We compare the effectiveness of our localization network trained
to re-use P2 features (\ie, the joint LP2 network),
and a network trained to do localization from scratch (\ie, as used in a sequential setting).
In both cases, faster inference is achieved by shallower (fewer layers)
and narrower networks (fewer channels) used for the online \lidar{} embedding.
The reported inference time does not include the map embedding branch, which can be pre-computed offline. 
The largest model corresponds to an architecture
similar to the P2 rasterized HD map backbone (which is itself 
a smaller version of the P2 \lidar{} backbone),
while the faster and smaller models have fewer layers or fewer channels in
each layer. The four largest models have 11 convolutional layers and a factor of $C =
1/2^{0}, 1/2^{1}, 1/2^{3}, 1/2^{4}$ the number of channels as the largest
model. The smallest (fifth largest) has $C = 1/2^{4}$ and five layers rather
than 11, which corresponds to one layer around each of the three
pooling/upsampling stages followed by a final layer.
We observe that, while reducing model size leads to a small drop in matching accuracy, this
does not end up affecting motion planning, as shown in
Table~\ref{tab:motion-planning}, while at the
same time reducing the online embedding computation time four-fold.

Finally, Table~\ref{tab:deep-gill-baselines-vs-multi-task-comparison}
compares our proposed online \lidar{} embedding networks to the
state of the art. The U-Net-based approach from~\cite{deep-gill} was shown to 
outperform classic approaches like ICP-based localization, especially in 
challenging environments such as highways.
Our results show that the original
performance can already be matched with a much faster network architecture, 
while leveraging the perception feature maps allows even smaller models to
perform at the same level. All inference times are measured on an NVIDIA
RTX5000 GPU.

\ifarxiv
\paragraph{Discussion:}
An alternative approach to tackle the LP2 problem is
to train a network from scratch that balances both the localization and
perception-prediction terms.
We tried this, but found that it performed worse than fine-tuning the localization module \emph{a posteriori}.
Understanding why this phenomenon occurs and training joint LP2 networks from scratch
is a question that we leave for future work.
\fi


\section{Conclusion}%
\label{sec:conclusion}




While prior research in autonomous driving has explored either full end-to-end
learning or the joint study of tasks such as object detection and motion
forecasting, the task of localization has not received as much attention in the
context of perception and planning systems, in spite of the strong reliance
of self-driving vehicles on HD maps for these tasks.

In this paper, we studied how localization errors 
affect state-of-the-art
perception, prediction, and motion-planning systems. Our analysis showed that
while perception is robust to relatively small localization errors, motion
planning performance suffers more, especially in case of yaw errors, motivating
the need to detect and correct such issues.
We subsequently proposed a multi-task learning solution capable of jointly
localizing against an HD map while also performing object detection and motion
forecasting, and showed that localization errors can be successfully detected
and corrected in less than 2 ms of GPU time.


\ifarxiv
Our work suggests multiple areas for improvement that may be addressed in future work.
For example, it would be interesting to evaluate our proposed architecture in a closed-loop
setting with both reactive actors and counterfactual sensor inputs.
We would also like to better characterize the performance of a localizer trained
jointly with other autonomy components, including metrics like
its localization recovery rate, its localization accuracy when enhanced with a time-aware pose filter, and
its robustness to changes in the map 
(\eg, due to construction, road re-pavement, or seasonality).
Moreover, future work could benefit from a more thorough comparison with
classical localization approaches such as those based on LiDAR registration~\cite{yoneda2014lidar}.
Similarly, future work may further evaluate the motion planner by classifying
the incurred collisions according to their severity, which would enable a better
understanding of the types of errors that arise due to localization error.
Finally, an alternative to jointly learning LP2 networks is to train P2 modules
with imperfect localization as input akin to data augmentation, 
while also applying regularization to avoid pathological cases of P2 modules that simply ignore the HD map altogether.
These are all interesting areas of future work.
\fi

{\small
\bibliographystyle{ieee_fullname}
\bibliography{CVConferencesAbrv,refs}
}

\end{document}